\documentclass[]{ecai} 

\usepackage{subcaption}

\usepackage[absolute,overlay]{textpos}

\usepackage{graphicx}
\usepackage{latexsym}
\usepackage{color, soul}
\usepackage{booktabs}
\usepackage{multirow}
\usepackage{stfloats}
\usepackage{microtype}
\usepackage{amsmath, amssymb}
\usepackage{mathtools}
\usepackage{resizegather}
\usepackage{amsthm}
\newtheorem{thm}{Theorem}
\theoremstyle{remark}
\newtheorem*{remark}{Remark} 

\DeclarePairedDelimiter\abs{\lvert}{\rvert}%
\DeclarePairedDelimiter\norm{\lVert}{\rVert}%
\makeatletter
\let\oldabs\abs
\def\abs{\@ifstar{\oldabs}{\oldabs*}}
\let\oldnorm\norm
\def\norm{\@ifstar{\oldnorm}{\oldnorm*}}
\makeatother

\usepackage{algorithm}
\usepackage[noend]{algcompatible}
\usepackage{placeins}
\usepackage[nolist]{acronym}
\begin{acronym}
	\acro{fl}[FL]{Federated Learning}
	\acro{iid}[IID]{Independent and Identically Distributed}
	\acro{sgd}[SGD]{Stochastic Gradient Descent}
\end{acronym}
\usepackage[switch]{lineno}
\paperid{782}        %

\begin{document}

\begin{frontmatter}
	\begin{textblock*}{15cm}(6.5cm,2cm) 
		\Large
		European Conference on Artificial Intelligence (ECAI) 2024
	 \end{textblock*}
\title{Tackling Selfish Clients in Federated Learning}
\author[A]{\fnms{Andrea}~\snm{Augello}
\thanks{Corresponding Author. Email: andrea.augello01@unipa.it.}}
\author[C]{\fnms{Ashish}~\snm{Gupta}
}
\author[A]{\fnms{Giuseppe}~\snm{Lo Re}
} %
\author[B]{\fnms{Sajal K.}~\snm{Das}
} %

\address[A]{Department of Engineering, University of Palermo, Palermo, Italy}
\address[B]{Department of Computer Science, Missouri University of Science and Technology, Rolla, USA}
\address[C]{Department of Computer Science, BITS-Pilani, Dubai, UAE}

\begin{abstract}
\ac{fl} is a distributed machine learning paradigm facilitating participants to collaboratively train a model without revealing their local data. 
However, when \ac{fl} is deployed into the wild, some intelligent clients can deliberately deviate from the standard training process to make the global model inclined toward their local model, thereby prioritizing their local data distribution.  
We refer to this novel category of misbehaving clients as selfish.
In this paper, we propose a {\bf R}obust aggregation strategy for the {\bf \ac{fl}} server to mitigate the effect of {\bf Self}ishness (in short RFL-Self). 
RFL-Self incorporates an innovative method to recover (or estimate) the true updates of selfish clients from the received ones, leveraging robust statistics (median of norms) of the updates at every round. By including the recovered updates in aggregation, our strategy offers strong robustness against selfishness. 
Our experimental results, obtained on MNIST and CIFAR-10 datasets, demonstrate that just 2\% of clients behaving selfishly can decrease the accuracy by up to 36\%, and RFL-Self can mitigate that effect without degrading the global model performance.
 
\end{abstract}
\acresetall
\end{frontmatter}

\section{Introduction}
With an aim to train a Machine Learning (ML) model in a privacy-preserving manner, the researchers in~\cite{mcmahan2016federated} introduced \ac{fl} framework, and since then it has drawn interest from the ML community.
Being a distributed learning paradigm, \ac{fl} enables each participant (or client) to train the model locally and send only model weights (parameters) to a central server for aggregation, thereby enabling learning from distributed heterogeneous devices without sharing their sensitive data.
\ac{fl} has been adopted in many contexts, such as, medical records management~\cite{rieke2020future}, activity recognition~\cite{concone2022federated}, and smart homes~\cite{gupta2022fedar}. 
Despite the advantages of \ac{fl}, the lack of oversight over the training process can have serious implications.
For instance, a client may deviate from the normal training~\cite{li2019abnormal}, negatively affecting the underlying model.
The deviation may be caused by a malicious client who wants to disrupt training~\cite{jere2020taxonomy}, or by a normal client who has insufficient resources~\cite{gupta_luo_ngo_das_2022,reisizadeh2022straggler}.
Malicious clients may introduce noise in the model to prevent convergence, as in byzantine attacks, or optimize for a secondary objective, e.g., a backdoor~\cite{wang2020attack}.
As the server lacks control over the clients, preventing their harmful actions can be challenging.
The issue of malicious or adversarial clients is a very active field of \ac{fl}. Broadly, two common approaches exist: (i) detect and remove the clients deviating from normal behavior~\cite{li2019abnormal,zhang2022fldetector} and (ii) mitigate the impact of the misbehaving clients via robust aggregation~\cite{awan2021contra,ma2022shieldfl}.
For instance, the server in~\cite{fung2020limitations} excluded the contribution of colluding malicious clients who send similar updates. 
Robust aggregation strategies, employing statistics such as median and trimmed estimators, also exhibit good performance in the presence of byzantine clients without needing to identify them~\cite{pmlr-v80-mhamdi18a}.
As long as the percentage of malicious clients stays below a certain threshold, the defense mechanism in~\cite{cao2021provably} can theoretically provide the correct classification.
\begin{figure}[tb]
	\centering
	\includegraphics[width=1\linewidth]{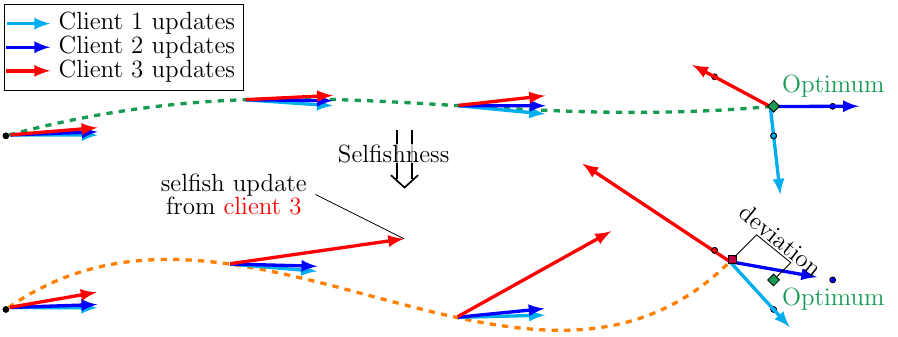}
	\caption{An example to show how selfish clients try to alter their updates to steer the global model toward their local optima.}
	\label{fig:selfish_effect}
\vspace{0.2in}
\end{figure}

In this paper, we focus on a previously unexplored type of misbehaving client, the {\em selfish client}.
Unlike malicious clients who intend to compromise the model, a selfish client is interested in making the global model prioritize its local data distribution.
The resultant model may cause performance degradation for normal clients, especially in non-\ac{iid} settings, though it is not the goal of the selfish client.
Figure~\ref{fig:selfish_effect} depicts an example of how selfishness deviates the global model from reaching the optimum.
Typically, each participating client sends updates toward its local optimum, which the \ac{fl} server aggregates to obtain a global optimum.
 However, a client behaving selfishly (shown by the `red' arrows) sends updates that differ in magnitude and direction from the ones it would send usually.

 Selfish updates cannot be directly included in the aggregation as that would lead to a deviated global model.
However, unlike malicious clients, selfish clients do not have any malign intent.
Their updates cannot be completely discarded as doing so would disregard their valuable contributions.
This also discourages the adoption of prior defense methods (against malicious clients)~\cite{jere2020taxonomy,gupta_luo_ngo_das_2022,wang2020attack} to deal with the selfishness issue.
To the best of our knowledge, we are the first to tackle selfish clients in FL.
We make the following major contributions: 
\begin{itemize}
    \item We introduce a novel concept of selfishness, caused by so-called selfish clients, in \ac{fl}.
    Under non-\ac{iid} settings, the presence of selfish clients is unavoidable as some intelligent clients can notice the difference between global and local optimum, and can try to exploit it in their favor. 
    \item We propose an innovative {\bf R}obust aggregation strategy that enables the {\bf \ac{fl}} server to mitigate the {\bf Self}ishness effect from the training process (RFL-Self) without affecting the overall performance of the global model. In RFL-Self, the server aims to recover the true updates from the received selfish updates before aggregation. 
	\item Through extensive experiments on two widely used datasets for \ac{fl} benchmarking, MNIST and CIFAR-10, we analyze the detrimental impact of selfish clients and demonstrate the effectiveness of RFL-Self for varying levels of selfishness and number of selfish clients.  
\end{itemize} 

The paper is organized as follows.
We first describe our selfish client behavior model and objective. 
Then, we introduce a selfish behavior implementation, assess the impact of selfish clients on \ac{fl}, and present a mitigation method.
We then discuss experimental results and finally conclude the paper.

\section{Preliminaries and Problem Description}\label{sec:Methods}

In \ac{fl}, the goal of the server is to find an optimal global model $\mathbf{w}^*$ satisfying the optimization problem: 
\begin{equation}\label{eq:fl-opt}
	\mathbf{w}^* = \arg\min_{\mathbf{w}} \left\{
		F(\mathbf{w}) \triangleq \frac{1}{k}\sum_{i\in [k]} F_i(\mathbf{w})
	\right\}, 
\end{equation}
where $F_i(\mathbf{w})$ is the loss function of $i$-th client and $[k]$ is the set of all clients.
To solve this optimization problem, at each communication round $t$, each client $i$ computes a local update $\delta_i^t$ through \ac{sgd}.
The updates are then sent to the server which computes a global update $\delta_{[k]}^t = \frac{1}{k}\sum_{i\in[k]}\delta_i^t$ as the average of all the local updates and obtains a new global model as $\mathbf{w}^{t+1}= \mathbf{w}^t + \delta_{[k]}^t$ for next round $t+1$.
To avoid overburdening the notation, from now on we omit the round superscript when clear from context.

\subsection{Selfish Client}\label{subsec:Selfish-client}

The optimization objective in Eq.~\eqref{eq:fl-opt} aims to learn a globally optimal model rather than providing an optimal model for each client. 
In non-\ac{iid} settings, the local optima of each client might differ from the global one.
Some intelligent clients might notice this difference and be interested in obtaining a global model closer to their local optima.
We refer to these savvy clients as {\em selfish} in \ac{fl}.
Through their actions, the global model ceases to be optimal for the whole system. 

It is worth noting that using a personalized model for each client~\cite{tan_towards_2022} would ensure that each client obtains a locally optimal, for instance by training multiple models for different groups of similar clients~\cite{augello_2023_dcfl}.
However, since selfish clients try to influence the global model at inference time, personalization techniques are unsuitable for their goals, as local fine-tuning has no impact on the performance of the global model.
Thus, in this work we focuses on the case where a single global model is trained for all clients, which is the most common \ac{fl} setup.

\noindent {\bf Selfish versus malicious clients:} The notion of selfish clients completely differs from the malicious ones (adversaries) that are well defined in the literature~\cite{blanco2021achieving}.
Selfish behavior does deviate from normal behavior, but the objective is not explicitly in contrast with the global objective.  
Selfish clients lack nefarious intent, such as engaging in model poisoning attacks or preventing model convergence, as malicious clients do.
Unlike backdoor attackers~\cite{wang2020attack}, selfish clients do not optimize the model for an additional objective different from the main task. 
The term ``selfish'' has been used in the context of \ac{fl} in a complementary way in~\cite{luo2021rethinking}, where it refers to a server that strives to favor a subset of clients over others, which is the opposite problem to the one addressed in this work.
Additionally, in certain works, ``selfish'' denotes typical malicious clients~\cite{youssef2022impact}.

Further, in communication networks, selfish clients are those wishing to reap benefits from a system without contributing~\cite{rahim2018vehicular}.
In \ac{fl}, these clients are commonly defined as ``free-riders''~\cite{pmlr-v130-fraboni21a} who wish to obtain a model without engaging in the training process, exhibiting a completely different behavior than the one addressed in this work.
A subset of researchers has utilized the term ``selfish'' to denote free-riders in their work~\cite{arisdakessian2023coalitional,sagduyu2022free}. However, the problem tackled in these studies is unrelated to the one addressed in this paper. As far as we are aware, this study is the first to address a similar issue.

\subsection{Problem Description}
A selfish client $s \in [k]$ mainly aims to craft local updates such that, after aggregating the local updates from all the clients, the global model incurs a lower loss on its local data. The client $s$ can formulate the objective for their local update as
\begin{equation}\label{eq:selfish_objective}
	\hat\delta_s =  \arg\min_{\delta} \left\{ F_s\left(\hat{\mathbf{w}}\triangleq \mathbf{w}  +
		\frac{\delta_{[k]\setminus \{s\}}+\delta}{k} \right) \right\},
\end{equation}
where the input to the objective function $F_s(\cdot)$ is not the global weights $\mathbf{w}$ but the crafted weights $\hat{\mathbf{w}}$ and two additional additive terms.
The first term $\frac{\delta_{[k]\setminus \{s\}}}{k}$ involves knowing the sum of all the local updates of the rest of the clients, which is generally not known. 
The first term $\frac{\delta}{k}$ can be easily computed for any given $k$, but determining the optimal $\delta$ is not straightforward.

Though the selfish clients do not intentionally aim to pollute the training process, their actions are not innocent. 
When clients have non-\ac{iid} data, the updates from selfish clients can considerably diverge the global model from the global optimum, causing performance degradation for normal clients.

To this end, the problem at hand is two-fold: 
\begin{enumerate}
	\item {\em From selfish client perspective:} solving the optimization problem in Eq.~\eqref{eq:selfish_objective} given the global model weights $\mathbf{w}$ and $k$. Similar to~\cite{bagdasaryan2020backdoor}, we assume clients know the total number of \ac{fl} participants ($k$).
    \item {\em From server perspective:} Alleviating the effect of selfishness on the global model through {\em robust aggregation}. Since selfish behavior differs from existing works on outliers and malicious clients, separate analysis and countermeasures are warranted.
\end{enumerate}

Due to the non-poisonous intent of the selfish clients, we can neither totally omit their updates from the aggregation nor include them directly, which makes the problem interesting and challenging compared to dealing with malicious clients.

\section{Proposed Approach}\label{sec:proposed}
To {introduce selfish clients in the} \ac{fl} context, we first propose a novel way to estimate selfish model updates and then present a robust aggregation method to mitigate the effect of selfishness on the global model. This work considers two types of clients: normal and selfish. Each client has heterogeneous data and selfish clients do not collude.

\subsection{Client Side: Computing Selfish Updates}\label{subsec:client}
To improve accuracy on its local data distribution, a selfish client needs to reduce the distance between global model updates $\delta_{[k]}$ (see Eq.~\eqref{eq:selfish_objective}) and the local model updates $\delta_s$ trained on local data.
To do so, at each round, the selfish client estimates $\hat{\delta}_s$ using the received global model $\mathbf{w}$ and prior knowledge of $k$. 
It is worth noting that even if $k$ is not known, the selfish client can still easily craft its updates approximating $k$, e.g., by jointly estimating it with ${\delta}_{[k]\setminus \{s\}}$.
This alternative is further discussed in the Appendix (Algorithm~\ref{alg:Estimation-of-the-number-of-FL-participants}).

\begin{figure}[tb]
	\centering
	\begin{subfigure}{0.4\linewidth}
		\includegraphics[width=\linewidth]{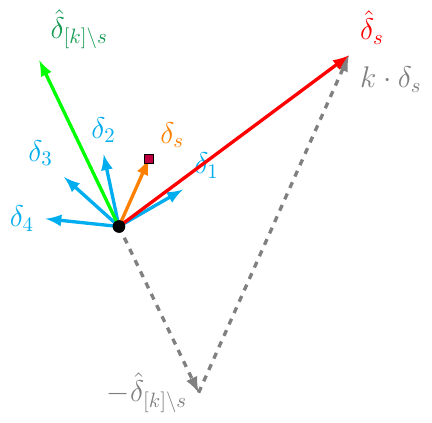}
		\caption{}
	\end{subfigure}
	\begin{subfigure}{0.55\linewidth}
		\includegraphics[width=\linewidth]{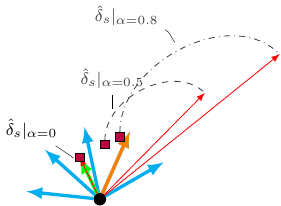}
		\caption{}
	\end{subfigure}
    \vspace{0.1in}
	\caption{
		(a) A visualization for how a selfish client can estimate $\hat{\delta}_s$ through model replacement. 
	(b) Effect of $\alpha$ on the aggregated model and selfish update vector of a single selfish client. The aggregated model $\hat{\mathbf{w}}$ is obtained when $\hat{\delta}_s$ is included in the aggregation.
 }
     \vspace{0.2in}
	\label{fig:alpha_effect_drawing}
\end{figure}

We propose an innovative strategy for the selfish client to estimate $\hat{\delta}_s$ to decrease the distance between the aggregated update $\delta_{[k]}$ and $\delta_s$. 
The idea behind our strategy is partly inspired by the model replacement attacks~\cite{bhagoji2019analyzing}, in which a malicious client attempts to replace the global model with its local model to neutralize the effect of other participants. 
The selfish client sends $\hat{\delta}_s$ in place of its true update $\delta_s$, however, unlike model replacement attacks, it does not have malicious intent.
A key distinction between model replacement attacks and selfish updates lies in their objectives. Model replacement attacks aim to manipulate the model towards a backdoor or impede convergence. Conversely, selfish updates seek to enhance the model performance based on (at least) some clients' data. In a perfect IID setting, we expect selfish updates to coincide with the true update, making the model converge to the same correct solution, unlike model replacement attacks.
Part (a) of Figure~\ref{fig:alpha_effect_drawing} depicts how the selfish client's update $\hat{\delta}_s$ can deviate from the updates sent by normal clients. 
By sending $\hat\delta_s$, the training advances toward a global model closer to the selfish client's local optimum.

\noindent {\bf Challenge:}
To be effective, the selfish client needs to know the update $\delta_{[k]\setminus \{s\}}$ resulting from the sum of the other clients' updates. Obtaining this knowledge is generally not possible in a standard \ac{fl} framework due to privacy concerns. Since we assume that the server is trustworthy and it communicates with clients via a secure channel, selfish clients cannot acquire $\hat\delta_{[k]\setminus \{s\}}$, thereby making it challenging to solve Eq.~\eqref{eq:selfish_objective}

To deal with the above challenge, a selfish client can instead estimate the average normal update $\bar\delta_{[k]\setminus \{s\}}$ assuming that it does not change much between two consecutive communication rounds.
Under this assumption, at round $t$, the selfish client uses the history of the global updates to compute an estimate of the average update $\bar\delta_{[k]\setminus \{s\}}$ as
\begin{equation}\label{eq:avg_update}
	\bar\delta_{[k]\setminus \{s\}}^t =
         \frac{k\cdot (\mathbf{w}^t - \mathbf{w}^{t-1}) - \hat\delta_s^{t-1}}{k-1} =
         \frac{{\delta}_{[k]\setminus \{s\}}^{t-1}}{k-1},
\end{equation}
where for $t=1$, $\hat{\delta}^{0}_{s} = \delta^{0}_{s}$. 
This formulation is valid for neural network-based models, tree-based~\cite{li2023fedtree} models were not considered in this work.
If multiple selfish clients are present, we assume that they are independent and each of them estimates their updates independently.
After obtaining this estimate, the selfish client aims to bring the subsequent global update closer to its local one. 
To achieve this, the client might send $\hat\delta_s= k \cdot \delta_s - (k-1) \cdot \bar\delta_{[k]\setminus \{s\}}$ as an update to the server, completely replacing the global model, unintentionally resulting in malicious behavior.
If the selfish client were to pursue this strategy, however, it would have no benefit in participating in the \ac{fl} system.
Thus, we introduce a selfishness parameter $\alpha \in [0,1]$ to control the fraction of the global update that is replaced with $\hat{\delta}_s$ in every round. 
With a constant $\alpha$ across all rounds, the selfish client estimates its updates as
\begin{equation}\label{eq:selfishness}
	\begin{aligned}
		\hat\delta_{s} & = \alpha\left(k\delta_s - ({k-1}){\bar\delta_{[k]\setminus \{s\}}}\right)
	+(1-\alpha)\bar\delta_{[k]\setminus \{s\}} \\
					   &= \alpha k(\delta_s - \bar\delta_{[k]\setminus \{s\}}) + \bar\delta_{[k]\setminus \{s\}}.
	\end{aligned}
\end{equation}

When multiple selfish clients are present, we assume that they are independent and do not collude with each other.
Each selfish client tries to optimize Eq.~\eqref{eq:selfish_objective} independently using Eq.~\eqref{eq:selfishness}, irrespective of the presence of other selfish clients.
Thus, the formulation holds with an arbitrary number of selfish clients, and  we expect them to compete to increase their effect on the global model and try to cancel each other's effect.

A trivial alternative strategy that could be employed by a selfish client is to simply upscale its update.
This strategy is equivalent to the presented one if $\delta_s = -\bar\delta_{[k]\setminus \{s\}}$, which is a special case of selfish update.
In an adversarial setting, where the misbehaving client aims to harm the model convergence, upscaling can be equally effective, but the same is not true in our setting.
In more realistic circumstances, upscaling can steer the model in a suboptimal direction. 

\noindent \textbf{Effect of parameter $\alpha$:}
With $\alpha=0$, the estimated update is equivalent to the other clients' updates, whereas, with $\alpha=1$, the selfish client would overwrite the whole global update with its local update.
In less extreme cases, selfishness can be regarded as the client trying to increase its effect during the aggregation process.
If $ \alpha= \frac{1}{k} $, then the update will simply be $ \hat{\delta}_s|_{\alpha=\frac{1}{k}} = \delta_s $.
On the other hand, $ \hat{\delta}_s|_{\alpha=\frac{k-1}{k}}=(k-1)\delta_s -(k-2)\bar{\delta}_{[k]\setminus \{s\}} $, which affects the global model update as if all the clients but one have an update equivalent to the selfish client and the remaining client has the update equivalent to the average update of the normal clients.
More generally, each $\frac{1}{k}$ increment of $\alpha$ increases the magnitude of the selfishness effect as if one more normal update $\delta_s$ from the selfish client were added and decreases the aggregated effect by one normal client.  A visual representation of the effect of the parameter $\alpha$ on the selfish update vector and the aggregated model $\hat{\mathbf{w}}$ is shown in part (b) of Figure~\ref{fig:alpha_effect_drawing}.

\begin{figure}[ht]
	\centering
	\includegraphics[width=0.95\linewidth, height=4.4cm]{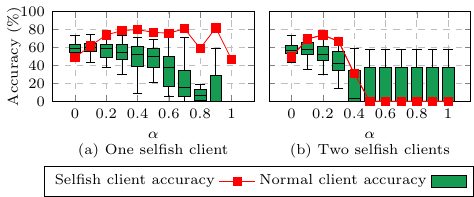}
	\caption{Illustrating the impact of selfishness on the test accuracy of the global model using CIFAR-10 dataset.} 
	\label{fig:5-50clients_no_mitigation_cifar}
\vspace{0.2in}
\end{figure}

\noindent \textbf{Impact of selfish updates:} We perform initial experiments using the CIFAR-10 dataset~\cite{krizhevsky2009learning} to demonstrate the impact of non-colluding selfish client(s) on the test accuracy of the global model with varying $\alpha$. 
The results for two different \ac{fl} setups (with $k=50$) are reported in Figure~\ref{fig:5-50clients_no_mitigation_cifar}.
The clients' data distributions are non-\ac{iid} with class partitioning so that each client has only the data of two randomly selected classes and every class of the dataset is assigned to at least one client.
The global model is tested on each client's test set to evaluate its performance on their data distribution.
More details on the experimental setup are given in Section~\ref{sec:Experimental-evaluation}.
It is easy to notice that increasing the selfishness (i.e., increasing $\alpha$) improves the accuracy for the selfish clients if there is only one in the \ac{fl} setup.
However, if a second selfish client appears, this is only true up to $\alpha=0.
3$. In the presence of two selfish clients with any same $\alpha$, each of them tries to cancel out the other's updates, making the global updates unbounded in magnitude, and consequently, harming the convergence of the training process, which can be observed in part (b) of Figure~\ref{fig:5-50clients_no_mitigation_cifar} for $\alpha\geq0.4$.   
In both cases, the performance for normal clients is harmed by the selfish clients.
With 3 or more selfish clients without a mitigation strategy convergence is not achieved.

\subsection{Server Side: RFL-Self}\label{subsec:Selfishness-mitigation}
When selfish clients participate in \ac{fl}, the objective of the server is to identify them before aggregation to mitigate their impact on the global model. 
However, even if these clients are detected correctly, they should not be completely excluded from the aggregation. 
Doing so leads to a model that performs poorly for the users (not available during training) having data distribution similar to the selfish clients. 
This can be observed in Figure~\ref{fig:5-50clients_no_mitigation_cifar} where with $\alpha=0$ (no selfish client contribution), the performance of the global model decreases for the selfish client's data distribution.
The only viable solution for the selfishness issue is estimating the true updates of selfish clients and using them in aggregation instead of the ones received from selfish clients.
This section proposes a robust aggregation strategy, RFL-Self, to remove the effect of selfishness from the training process.
Each round, RFL-Self identifies suspected selfish clients and attempts to recover their true update.

\subsubsection{Identifying Selfish Clients}
Given a set of received updates from all the clients, RFL-Self first computes the $L_2$ norm of each update and finds a median norm $\mathcal{N}_{med}$. Since a selfish client aims to increase its influence on the global model to deviate the training toward its local model, it is obvious that selfish updates are larger in magnitude than those from other clients. 
This intuition, confirmed by Theorem~\ref{thm:selfish_norm}, suggests that for any client $i$, if $\norm{\delta_i}^2 > \mathcal{N}_{med}$ holds, then client $i$ might be selfish.
In this case, the received update $\delta_i$ would be the crafted $\hat\delta_i$.  
This detection strategy is intentionally crafted to be simple and lightweight, but it is fully supported by the theoretical analysis in Theorem~\ref{thm:selfish_norm}, and the experimental results confirm its effectiveness.

\begin{figure}[ht]
	\centering
	\includegraphics[width=0.95\linewidth, height=4.3cm]{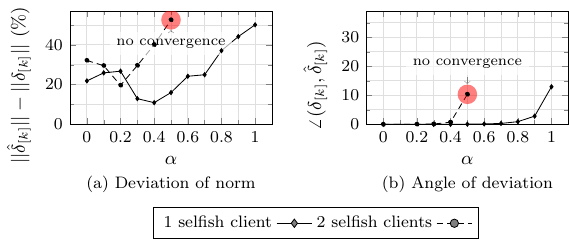}
	\caption{Deviation of the global update when all clients are normal to the one with selfish clients (CIFAR-10 dataset).}
	\label{fig:resilience}
\vspace{0.2in}
\end{figure}

Figure~\ref{fig:resilience} visually demonstrates the deviation of the global update $\delta_{[k]}$ caused by selfish updates. It showcases the disparity in norm and angle between the global update in scenarios where all clients are normal and where selfish clients are also present.
It can be observed that part (a) of Figure~\ref{fig:resilience} shows a U-shaped dip.
This phenomenon is caused by the magnitude of $\hat\delta_s + \delta_{[k]\setminus \{s\}}$ not being equal to the magnitude of $\delta_{[k]}$, as also illustrated in part (b) of Figure~\ref{fig:alpha_effect_drawing}.
Even a single selfish client can severely affect global convergence, and just one more selfish participant may lead to a {\em no convergence} situation. Thus, recovering the true updates of the selfish clients becomes crucial.   

\subsubsection{Recovering True Updates of Selfish Clients}
Upon successful identification of the selfish clients, 
there are three possible ways in which the server can deal with them:
\begin{itemize}
	\item {\em Drop}: exclude suspicious updates from the aggregation.
	\item {\em Mitigate}: reduce the impact of suspicious updates.
	\item {\em Recover}: try and recover the genuine update.
\end{itemize}

The exclusion of selfish updates is not desirable as these updates are not malicious and help improve the generalization of the model.
Moreover, normal clients might be wrongly identified as selfish in some rounds.
Thus we focus on recovery to preserve valuable contributions.
This approach is unique to our work as it does not apply to malicious clients.

RFL-Self uses the received selfish updates $\hat\delta_s$ to compute an estimate $\delta_s'$ of the true update $\delta_s$ and uses the estimate in the aggregation process.
The server uses a convex combination of the received update $\hat\delta_s$ of each identified selfish client $s$ and the marginal median update $\delta_{med}$ to obtain $\delta_s'$ as 

\begin{equation}\label{eq:mitigation}
	\delta_s' = \beta\hat\delta_{s} + (1-\beta)\delta_{med} , 
\end{equation}
where the parameter $\beta$ determines the convex combination and ensures that $||\delta_s'||$ is the same as the median of the norms of all the received updates $\mathcal{N}_{med}$. Thus, $\beta$ can be computed by solving the equation
\begin{align}\label{eq:beta}
    ||\beta\hat\delta_s +(1-\beta)\delta_{med}|| = \mathcal{N}_{med}.
\end{align}
Through the convex combination expressed in Eq.~\eqref{eq:mitigation}, the potentially selfish update is scaled down and rotated toward the median update, obtaining a recovered version of the true update $\delta_s$. 
Later in this section, we provide a theoretical analysis of the soundness of the recovery process and bounds on the possible reconstruction error.

\subsubsection{Robust Aggregation}
Finally, by replacing the received updates of the set of suspected selfish clients $\mathcal{S}$ with the recovered ones, the server aggregates the updates as expressed below:

\begin{equation}
    \delta_{[k]} = \frac{1}{k} \left(\sum_{i \in [k] \setminus \mathcal{S}}\delta_i + \sum_{j \in \mathcal{S}} \delta_j' \right).
\end{equation}
Later, the global model for the next round is computed as $\mathbf{w}= \mathbf{w} + \delta_{[k]}$.
By substantially mitigating the effect of selfishness, our aggregation process offers a robust \ac{fl} framework against the selfish participants while strategically utilizing their updates to achieve a more generalized global model.   

Algorithm~\ref{alg:mitigation} summarizes all the steps of the RFL-Self.
It is worth noting that RFL-Self differs from the standard aggregation method (FedAvg~\cite{mcmahan2016federated}) only at the server, thus it can be implemented as a drop-in replacement for FedAvg transparently from the clients without incurring any additional communication overhead.

On the server side, the overhead for computing the median of norms is $O(kd + k\log k)$, where $d$ is the number of model parameters. 
This complexity is negligible compared to $O(dk\log k)$ required for the median computation.
Thus, RFL-Self has asymptotically the same complexity as using the median as a robust aggregation mechanism and requires no additional communication.

\begin{algorithm}[tb]
	\caption{RFL-Self algorithm}
	\label{alg:mitigation}
	\begin{algorithmic}[1]
		\renewcommand{\algorithmicrequire}{\textbf{Input:}}
		\renewcommand{\algorithmicensure}{\textbf{Output:}}
		\REQUIRE A set of $k$ clients
		\ENSURE A trained global model with weights $\mathbf{w}$
		
		\STATE $\mathbf{w} \gets $ random initialization
		\STATE Send $\mathbf{w}$ to all the clients
		\FOR{{\em each communication round}}
			\STATE Receive  updates $\{\delta_i, \ \forall i \in [k]\}$
			\STATE Compute  $\pmb{\mathcal{N}} \gets \{||\delta_i||, \,\forall i \in [k]\}$
			\STATE Compute a median norm $\mathcal{N}_{med}$ of $\pmb{\mathcal{N}}$
			\STATE $\delta_{med} \gets \textsc{MarginalMedian}(\{\delta_i, \,\forall i \in [k]\})$
            \STATE $\mathcal{S} \gets \{\}$
            \FOR{$each \ client \ i \in [k]$}
                \IF{$||\delta_i|| > \mathcal{N}_{med}$}
                    \STATE $\mathcal{S} \gets \textsc{Append}(i)$
                \ENDIF
            \ENDFOR
			\FOR{$each \ client \ i \in \mathcal{S}$}
    			\STATE Obtain $\beta$ by solving Eq.~\eqref{eq:beta} 
    			\STATE Estimate $\delta_i'$ using Eq.~\eqref{eq:mitigation}
			\ENDFOR

			\STATE Update $\mathbf{w} \gets \mathbf{w} +\frac{1}{k} (\sum_{i \in [k] \setminus \mathcal{S}}\delta_i + \sum_{j \in \mathcal{S}} \delta_j' )$

			\STATE Send the updated global model $\mathbf{w}$ to all the clients
		\ENDFOR
		 \STATE \textbf{return} $\mathbf{w}$

	\end{algorithmic}
\end{algorithm}

\subsubsection{Theoretical Guarantees}
This section presents rigorous theoretical insights about the RFL-Self, focusing on the soundness of the detection and recovery mechanisms.
\begin{thm}\label{thm:selfish_norm}
If the true update is similar in magnitude to the average update of the normal clients, then an effective estimated update of a selfish client is always larger in magnitude than the true update.
\end{thm}
\begin{proof}
Let us assume that there exists some $\alpha$ such that $\oldnorm{\hat\delta_s}^2 = \norm{\delta_{s}}^2$, making the norm of the estimated update indistinguishable from the true update.
Such values of $\alpha$ can be determined by using Eq.~\eqref{eq:selfishness}.
 In the norm space, this condition can be written as 
{
\begin{gather}
	\scriptstyle \nonumber
	\hspace{-0ex} \norm{\alpha k \delta_s +  (1- \alpha k)\bar \delta_{[k]\setminus \{s\}}}^2 = \norm{\delta_s}^2,\\ 
	\scriptstyle \nonumber
	\hspace{-0ex} (\alpha k\norm{\delta_s})^2+2\alpha k(1-\alpha k)\langle \delta_s,  \bar\delta_{[k]\setminus \{s\}}\rangle+(1-\alpha k)^2\norm{\bar\delta_{[k]\setminus \{s\}}}^2 = \norm{\delta_s}^2,
	\shortintertext{\hspace{-0em}\normalsize where $\langle \cdot, \cdot \rangle$ is the inner product. After solving the above, we get }
	\scriptstyle
	\hspace{-1ex} \alpha=\frac{1}{k} \quad \text{\normalsize and } \quad \alpha = \frac{\norm{\bar\delta_{[k]\setminus \{s\}}}^2-\norm{\delta_s}^2}{k(\norm{\delta_s}^2+2\langle\delta_s, \bar\delta_{[k]\setminus \{s\}}\rangle+\oldnorm{\bar\delta_{[k]\setminus \{s\}}}^2)}\label{proof1}
\end{gather}
}
For $\alpha=\frac{1}{k}$ the $\hat{\delta}_s = \delta_s$, i.e., the estimated update coincides with the true update.
The other value for $\alpha$ in Eq.~\eqref{proof1}, if $\oldnorm{\delta_s}^2\simeq\oldnorm{\bar\delta_{[k]\setminus \{s\}}}^2$, corresponds to $\alpha\simeq0$, i.e. $\hat{\delta}_s\simeq\bar\delta_{[k]\setminus \{s\}}$.
Thus, the only way to avoid updates with a larger magnitude (larger norm) than the true ones, is by not behaving selfishly, while effective selfish updates are always larger in magnitude than the true updates.
\end{proof}
\begin{remark}
    Theorem~\ref{thm:selfish_norm} does not ensure that all clients exhibiting larger update norms are necessarily acting selfishly.
	Consequently, RLF-Self can incur false positives.
	This aspect is taken into account in the recovery process, ensuring that the simplicity of the detection mechanism does not undermine performance.
\end{remark}

\begin{figure}[ht]
	\centering
	\begin{subfigure}[b]{0.44\linewidth}
		\includegraphics[width=\linewidth]{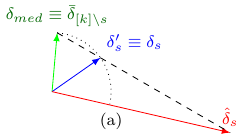}
	\end{subfigure}
	\begin{subfigure}[b]{0.44\linewidth}
		\includegraphics[width=\linewidth]{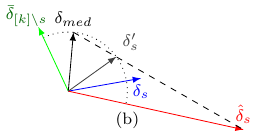}
	\end{subfigure}
	\caption{Distance of the recovered update $\delta_s'$ to $\delta_s$: (a) when the conditions 1-3 are met, and (b) when the three conditions do not hold.}
     \vspace{0.2in}
	\label{fig:mitigation_perfect}
\end{figure}

\noindent \textbf{Recovery error:}
The convex combination in Eq.~\eqref{eq:mitigation} reduces the effect of the selfish update steering it towards the normal clients' updates.
More formally, we analyze the recovery error of RFL-Self under the following conditions:
\begin{itemize}
	\item {\em Condition 1:} The average update $\delta_{[k]\setminus \{s\}}$ (excluding selfish updates) is similar between any two consecutive rounds. 
	\item {\em Condition 2:} The median update $\delta_{med}$ is a good estimator of the mean update of all but the selfish clients. 
	\item {\em Condition 3:} The true update by a selfish client $\delta_s$ is close in magnitude to $\mathcal{N}_{med}$. 
\end{itemize}
If the first two conditions hold, then 
$ \bar\delta_{[k]\setminus \{s\}} \simeq \delta_{med}$ and from Eqs.~(\ref{eq:selfishness}) and~(\ref{eq:mitigation}) we get
\begin{align}\label{eq:estimate}\nonumber
    \delta_s' & \simeq \beta[\alpha k (\delta_s - \bar\delta_{[k]\setminus \{s\}}) + \bar\delta_{[k]\setminus \{s\}}] + (1-\beta)\delta_{med} \\ 
    &\simeq \beta \alpha k \delta_s + (1-\beta\alpha k)\bar\delta_{[k]\setminus \{s\}}.
\end{align}
This recovered update $\delta_s'$ coincides with $\delta_s$ for $ \beta = 1/(\alpha k)$. The server, however, does not know  of $\alpha$, hence, it chooses $\beta$ 
by solving $ ||\delta_s'|| = \mathcal{N}_{med}$, as mentioned in Eq.~\eqref{eq:beta}. If all three conditions hold, then choosing $\beta$ using Eq.~\eqref{eq:beta} would provide $ \beta \simeq 1/(\alpha k)$, thereby taking $\delta_s'$ quite close to $\delta_s$.
Part (a) of Figure~\ref{fig:mitigation_perfect} shows the effectiveness of RFL-Self on recovery of the selfish update $\delta_s'$ if the three conditions are met. 

Even if all the conditions are not satisfied perfectly, the RFL-Self can still provide a good approximation. For example, dropping Condition 2, we get 
$ \delta_s' = \beta \alpha k \delta_s + (\delta_{med}-\beta\alpha k\bar\delta_{[k]\setminus \{s\}}) + \beta(\bar\delta_{[k]\setminus \{s\}}-\delta_{med})$, which differs by $ (1-\beta)(\delta_{med} - \bar\delta_{[k]\setminus \{s\}})$ from the one in Eq.~\eqref{eq:estimate}. The norm of this difference (error) is upper bounded by $ \sqrt{\mathrm{Tr}(\mathrm{var}(\delta))} $~\cite{garver1932concerning}.
Similarly, dropping  Condition 3, the estimated value will be closer to the median update than the original one, with a maximum error of $||\delta_s-\delta_{med}||$. 
Part (b) of Figure~\ref{fig:mitigation_perfect} shows the difference between the recovered update $ \delta_s'$ and the original update $ \delta_s $ when the three conditions do not hold.
It is worth mentioning that a trivial mitigation strategy, such as downscaling selfish updates by a factor $\beta$ before aggregation, does not provide the same guarantees.
Downscaling yields $ \delta_s' = \frac{\hat\delta_s}{\beta} = \frac{\alpha k}{\beta}\delta_s + \frac{1-\alpha k}{\beta} \bar\delta_{[k]\setminus \{s\}} $.
No matter how $\beta$ is chosen, the estimated update $\delta_s'$ cannot match the true update.
RFL-Self offers greater robustness by estimating $\delta_s'$ closer to the true update $\delta_s$, as also empirically demonstrated in Section~\ref{sec:Experimental-evaluation}.
\begin{thm}\label{thm:max-error}
    The maximum error in the recovered aggregated update is bounded by $\frac{4+k}{4k}  \, \mathrm{Tr(var(}\delta))$.
\end{thm}
\begin{proof}
Owing to Theorem~\ref{thm:selfish_norm}, in the presence of a set $\mathcal{S}$ of suspected selfish clients, the expected error is: 

{
\begin{align}\nonumber
	\scriptstyle
	\mathbb{E} \norm{\delta_{[k]}-\delta'_{[k]}}^2 &=   
	\scriptstyle
	\frac{1}{k^2}\mathbb{E}\norm{\sum_{i\in[k]}\delta_i - \left(\sum_{i\in [k]\setminus \mathcal{S}}\delta_{i} + \sum_{j\in\mathcal{S}}\delta'_{j} \right)}^2 \\
 & \scriptstyle
 = \frac{1}{k^2}\mathbb{E}\norm{\sum_{i\in\mathcal{S}}\delta_i - \sum_{j\in\mathcal{S}}\delta'_{j}}^2
\end{align}
}
Since the RFL-Self leaves all $[k]\setminus \mathcal{S}$ updates unchanged, the error comes from the updates whose norm is larger than $\mathcal{N}_{med}$. The maximum error induced by recovering these updates is obtained by using the median update instead

{
\begin{align}\nonumber
	\scriptstyle
		\frac{1}{k^2}\mathbb{E}\norm{\sum_{i\in\mathcal{S}}\delta_i - \sum_{j\in\mathcal{S}}\delta'_{j}}^2 
	&\scriptstyle
		\leq  \frac{1}{k^2}\mathbb{E}\norm{\sum_{i\in\mathcal{S}}
			\left( \delta_{i}-\delta_{med} \right) }^2 \\ \nonumber
	&\scriptstyle
		 =  \frac{1}{k^2}\mathbb{E}\norm{\sum_{i\in\mathcal{S}}
		\left( \delta_{i}-\bar\delta_{[k]} - \delta_{med}+\bar\delta_{[k]}\right) }^2 
		\scriptstyle
		\shortintertext{\normalsize Using Jensen's inequality~\cite{jensen1906fonctions}} \nonumber
		\leq 
	&\scriptstyle
	\frac{1}{k^2}  \mathbb{E}  \norm{\sum_{i\in\mathcal{S}}
			\left( \delta_{i} -\bar\delta_{[k]} \right) }^2  \!+  
			\frac{\abs{\mathcal{S}}^2}{k^2} \mathrm{Tr}(\mathrm{var}(\delta)) \\ 
	\scriptstyle \leq &\scriptstyle
	\frac{1}{k} \mathrm{Tr}  (\mathrm{var}(\delta)) 
		+ \frac{\abs{\mathcal{S}}^2}{k^2} \mathrm{Tr}(\mathrm{var}(\delta)) \\ 
	\scriptstyle \simeq &\scriptstyle
		 \frac{4+k}{4k}  \, \mathrm{Tr(var(}\delta)) \quad \text{\normalsize since } \abs{\mathcal{S}} \simeq \frac{k}{2}. \qedhere
\end{align}
}

\end{proof}

 \begin{remark}
  According to Theorem~\ref{thm:max-error}, the error does not depend on the number of selfish clients as long as they are insufficient to induce substantial bias into the median.
 \end{remark}

\section{Experimental Evaluation}\label{sec:Experimental-evaluation}
This section empirically analyzes the impact of selfish clients on \ac{fl} system performance and evaluates the effectiveness of our RFL-Self method. 
Under an image classification task, we use two benchmark datasets widely adopted to assess \ac{fl} algorithms, MNIST~\cite{deng2012mnist} and CIFAR-10~\cite{krizhevsky2009learning}, with 50 clients varying the level of selfishness $\alpha$ with up to 20\% of the clients being selfish. We chose to assess RFL-Self on these datasets due to their widespread use as benchmarks for FL and their varying complexities. Larger datasets like CIFAR-100 were not included because, in non-IID settings, the increased diversity in data distribution compels selfish clients to transmit even larger updates. Consequently, detecting selfish clients using our method becomes straightforward.
To simulate the non-\ac{iid} scenario, we partition the dataset so that each client has data for two randomly selected classes, which is the most challenging setting~\cite{li2022federated}, as few clients have overlapping classes.
The tested non-IID conditions represent one of the most challenging scenarios in FL. In less severe non-IID conditions, the influence of selfish clients is expected to be less pronounced. Additional experiments on a smaller FL setup involving 5 clients are detailed in the supplementary material.

\noindent \textbf{Experimental setup:}
For the MNIST dataset, we train a CNN with 2 convolutional layers followed by 2 fully connected layers.
For the CIFAR-10 dataset, the trained CNN has 3 convolutional layers instead.
Our investigation of \ac{fl} is focused solely on the perspective of selfishness, rather than maximizing accuracy. Therefore, these shallow models are suitable enough for our task.
The hyper-parameters are -- {\em optimizer}: \ac{sgd}, {\em batch size}: 128/256, and {\em learning rate}:  0.01 and 0.1 for MNIST and CIFAR-10 datasets, respectively. 
We train the models for 30 communication rounds, with five local epochs per round at each client, assuming full participation of all the clients in every round.
All the experiments are implemented in Python using a well-known library PyTorch 1.12.1~\cite{paszke2019pytorch} on a Windows 11 powered with an NVIDIA RTX A5000 GPU. 
The source code for this work is available at~\cite{augello_2024_12755207}.
\subsection{Impact of Selfish Clients on Model Performance}
\begin{figure}[tb]
	\centering
	\includegraphics[width=0.95\linewidth, height =4.2cm]{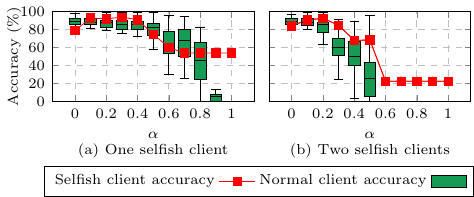}
	\caption{Test accuracy of the global model varying $\alpha$ with one and two selfish clients on the MNIST dataset.} 
     \vspace{0.1in}
	\label{fig:5-50clients_no_mitigation_mnist}
\end{figure}

At first, we carry out experiments to analyze the impact of selfish client(s) on the global model accuracy without using RFL-Self, and report results in Figure~\ref{fig:5-50clients_no_mitigation_mnist} for MNIST dataset. 
Similar results were discussed in Figure~\ref{fig:5-50clients_no_mitigation_cifar} for CIFAR-10 to ease the understanding of the context therein.
Though the mean test accuracy (shown by `green' box plots) across the normal clients keeps decreasing as the selfishness level $\alpha$ increases, it does not seem to favor the selfish clients. 
In part (b) of Figure~\ref{fig:5-50clients_no_mitigation_mnist}, for $\alpha>0.3$ the two selfish clients also start losing accuracy. It is interesting to observe that in the presence of two selfish clients, none of them gets much benefit, rather each seems to cancel out the other's updates, thus getting trapped into a no-win situation. Additionally, the selfishness causes a higher variance of the test accuracy of the global model across normal clients, thus decreasing the overall performance of the \ac{fl} system, and in turn, making the model more unpredictable and unfair to the normal participants.

\begin{figure}[tb]
	\centering
 \includegraphics[width=0.95\linewidth, height=3.8cm]{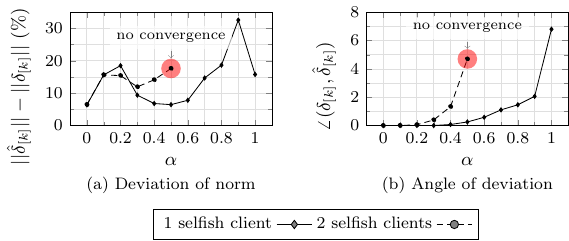}
	\caption{Deviation of the global update when all clients are normal to the one with selfish clients (MNIST dataset).}
     \vspace{0.2in}
	\label{fig:resilience_mnist}
\end{figure}

In addition, to analyze how selfishness affects the model convergence, we compute the deviation between the global update in the absence of selfish clients $\delta_{[k]}$ and the global update in the presence of selfish client(s) without applying RFL-Self $\hat\delta_{[k]}$.
The obtained results for MNIST dataset are reported in Figure~\ref{fig:resilience_mnist}. 
We also showed similar results for CIFAR-10 in Figure~\ref{fig:resilience} to establish the need for our aggregation method. 
An important point to notice here is that in the presence of two selfish clients, at 
$\alpha=0.5$,
the global model fails to converge though the deviation is not large. It is happening because each selfish client cancels out the selfishness component of the other, thereby the overall model weights keep fluctuating. 
 These results in the absence of any mitigation strategy further emphasize the necessity of a robust aggregation mechanism to handle selfish clients in \ac{fl} systems.

\subsection{Performance of RFL-Self}
Next, we evaluate the performance of RFL-Self in the presence of selfish clients and make a fair comparison with two standard strategies: median and downscaling. The median is also a robust aggregation strategy in which the median is used instead of the mean when aggregating, it is commonly used in \ac{fl} systems to deal with outliers and byzantine clients~\cite{yin2018byzantine}, and is representative of the ``drop'' strategy that completely excludes the updates of the suspected selfish clients from the aggregation process.
In the downscaling strategy, whenever a client is suspected to be selfish in a given round, instead of using Eq.~\eqref{eq:mitigation}, we scale each suspected update by ${\mathcal{N}_{med}}/{\oldnorm{\hat\delta_s}}$. 
The authors in~\cite{gupta_luo_ngo_das_2022} adopted downscaling to deal with unreliable clients, here we use it as a representative of the ``mitigate'' strategy.
While sophisticated malicious client detection mechanisms have been proposed in the literature, in our scenario, Theorem~\ref{thm:selfish_norm} proves that our criterion, albeit simple (with a low computational overhead), is guaranteed to be effective.
The main contribution on the defense side of the work lies in the original update recovery, not in the flagging mechanism.
Nevertheless, the proposed flagging mechanism can be substituted with different ones without affecting the recovery process.

\begin{figure}[tb]
	\centering
	\includegraphics[width=0.9\linewidth,height=6.6cm]{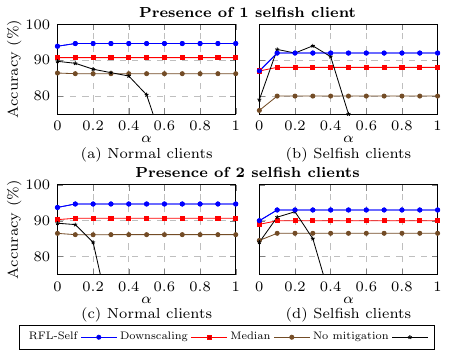}
	\caption{Performance of the robust aggregation strategies under an \ac{fl} setup with 50 clients using MNIST dataset.}
	\label{fig:mitigation_alpha_variable_mnist}
     \vspace{0.2in}
\end{figure}

\begin{figure}[tb]
	\centering
 \includegraphics[width=0.9\linewidth,height=6.6cm]{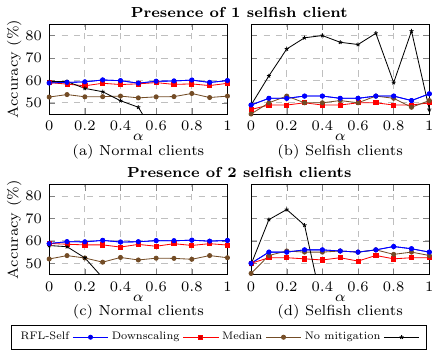}
	\caption{Performance of the robust aggregation strategies under an \ac{fl} setup with 50 clients using CIFAR-10 dataset.}
     \vspace{0.2in}
	\label{fig:mitigation_alpha_variable_cifar}
\end{figure}

We test all three strategies in the same settings as the selfish client impact assessment.
The test accuracy results for selfish and normal clients with varying $\alpha$ are reported in Figures~\ref{fig:mitigation_alpha_variable_mnist} and~\ref{fig:mitigation_alpha_variable_cifar}.

A quick observation from the results is that selfishness with $\alpha>0.4$ in FL can cause severe accuracy degradation for both normal and selfish clients if no robust aggregation has been used at the server. 
An ideal mitigation strategy should prevent accuracy reduction for normal clients without affecting selfish clients. In that sense, all the considered strategies seem to perform well by preventing selfish clients from causing drastic accuracy reductions. Additionally, the RFL-Self method outperforms the downscaling by $4\%\sim5\%$ and the median by $7\%\sim12\%$ on MNIST dataset for all $\alpha>0$, because our method avoids excessive penalties for selfish clients while still preventing them from gaining much on the accuracy. 
Interestingly, selfish clients never achieved more accuracy than normal ones regardless of the selfishness level. Another significant observation is that, even when selfishness levels are low, RFL-Self outperforms the ``No mitigation'' scenario. Even in the absence of selfish clients, the proposed method does not compromise the performance.

\subsection{Performance with More Selfish Clients}

Finally, we investigate how the performance of the considered strategies changes by increasing the number of selfish clients with $\alpha \in \{0.2, 0.3, 0.4\}$. We choose these values of $\alpha$ with the help of part (b) of Figure~\ref{fig:5-50clients_no_mitigation_cifar}, where at $\alpha=0.2$, the model achieves maximum accuracy even with two selfish clients and right after that it starts degrading and at $\alpha=0.4$, we notice a steep drop in accuracy. Table~\ref{tab:mitigation_selfish_variable} reports the results on the CIFAR-10 dataset under an \ac{fl} setup with 50 clients including the selfish clients (0\% to 20\%).
On the CIFAR-10 dataset, at $\alpha=0.3$ with $10\%$ selfish clients, the downscaling loses to our method by a substantial $4.2\%$ margin $(60.84-56.64)$.
Regardless of the percentage of selfish clients, RFL-Self outperforms both downscaling and median and maintains a fair balance between the accuracy of normal and selfish clients.        
Notably, RFL-Self outperforms the other strategies even in the absence of selfish clients, supporting that it does not harm the performance of normal clients.

\begin{table}
	\centering
	\caption{Performance of the global model and standard deviation under the different aggregation strategies on the CIFAR-10 dataset.}
	\resizebox{\linewidth}{!}%
 {	
	 \newcommand{\PM}{\ensuremath{\scriptstyle\pm}}
	\begin{tabular}{lrr|*{3}{r}|*{3}{r}*{3}{r}|*{3}{r}}
		\toprule
		 $\alpha$ & \multicolumn{2}{c|}{Selfish \%}& 
		  \multicolumn{3}{c|}{Normal clients accuracy}&
		 \multicolumn{3}{c}{Selfish clients accuracy}\\
		 & \multicolumn{2}{c|}{(\# clients)} & RFL-Self & Downscaling & Median &  RFL-Self & Downscaling & Median  \\ \midrule

		 \multirow{4}{*}{0.2}
		 &0  & (0)&   $\textbf{59.58}\PM6.54$ & $59.28\PM6.23$ & $52.42\PM7.11$ & -     & -     & -     \\
		 &5  & (3)&   $\textbf{59.11}\PM6.17$ & $57.83\PM6.31$ & $57.67\PM9.18$ & $\textbf{56.67}\PM1.35$ & $55.00\PM1.94$ & $54.53\PM3.57$\\
		 &10 & (5)&   $\textbf{57.78}\PM6.52$ & $56.91\PM7.04$ & $55.64\PM8.41$ & $\textbf{57.80}\PM2.40$ & $54.80\PM2.26$ & $56.60\PM7.74$\\ 
		 &20 & (10)&  $\textbf{56.92}\PM7.11$ & $55.95\PM8.72$ & $52.12\PM9.64$ & $\textbf{60.00}\PM6.13$ & $58.30\PM8.58$ & $55.00\PM8.26$\\ \midrule
                                                  
		 \multirow{4}{*}{0.3}
		 &0 & (0)&    $\textbf{59.58}\PM6.54$ & $59.28\PM6.23$ & $52.42\PM~7.11$ & -     & -     & -     \\
		 &5  & (3)&   $\textbf{59.64}\PM6.24$ & $58.13\PM6.65$ & $53.98\PM~9.89$ & $\textbf{55.33}\PM1.88$ & $53.00\PM2.08$ & $54.33\PM~4.73$\\
		 &10 & (5)&   $\textbf{60.84}\PM6.76$ & $56.64\PM6.22$ & $55.47\PM10.18$ & $\textbf{56.00}\PM2.12$ & $54.20\PM1.81$ & $56.20\PM~8.05$\\ 
		 &20 & (10)&  $\textbf{56.52}\PM6.28$ & $55.88\PM6.10$ & $52.80\PM~9.20$ & $\textbf{59.80}\PM6.88$ & $58.20\PM6.96$ & $56.20\PM10.19$\\\midrule
                                                  
		 \multirow{4}{*}{0.4}                    
		 &0  & (0)   &  $\textbf{59.58}\PM6.54$ & $59.28\PM6.23$ & $52.42\PM~7.11$ & -     & -     & -     \\
		 &5  & (3)   &  $\textbf{59.94}\PM6.16$ & $58.51\PM6.13$ & $53.15\PM~9.51$ & $\textbf{55.67}\PM2.49$ & $52.33\PM2.65$ & $54.33\PM4.16$ \\
		 &10 & (5)   &  $\textbf{60.49}\PM6.49$ & $58.91\PM8.73$ & $54.33\PM10.38$ & $\textbf{56.00}\PM2.23$ & $54.40\PM6.37$ & $49.60\PM7.33$ \\ 
		 &20 & (10)  &  $\textbf{60.08}\PM9.83$ & $59.05\PM9.09$ & $53.77\PM~9.01$ & $\textbf{58.70}\PM9.08$ & $58.30\PM9.23$ & $54.80\PM8.79$ \\ \bottomrule
	\end{tabular}      
}                     

	\label{tab:mitigation_selfish_variable}
\end{table}

Table~\ref{tab:mitigation_selfish_variable_mnist} reports the obtained results for MNIST dataset with 50 clients including the selfish clients (0\% to 20\%).
On a simple dataset like the MNIST, the normal clients can achieve more than $90\%$ accuracy via downscaling, closer to the performance of RFL-Self.

\begin{table}
	\centering
	\caption{Performance of the global model and standard deviation under the different aggregation strategies on the MNIST dataset.}
 \resizebox{\linewidth}{!}%
 {	
	\newcommand{\PM}{\ensuremath{\scriptstyle~\pm~}}
	\begin{tabular}{lrr|*{3}{r}|*{3}{r}*{3}{r}|*{3}{r}}
		\toprule
$\alpha$ & \multicolumn{2}{c|}{Selfish \%}& 
		  \multicolumn{3}{c|}{Normal clients accuracy}&
		 \multicolumn{3}{c}{Selfish clients accuracy}\\
		 & \multicolumn{2}{c|}{(\# clients)} & RFL-Self & Downscaling & Median &  RFL-Self & Downscaling & Median  \\ \midrule  

		 \multirow{4}{*}{0.2}
		 &0 & (0)&    $\textbf{90.98}\PM3.66$ & $90.66\PM3.82$ & $86.22\PM3.86$ &   -       &    -    &    - \\
		 &5 & (3)&    $\textbf{92.34}\PM3.55$ & $91.81\PM3.96$ & $88.40\PM3.83$ &   $\textbf{92.00}\PM2.11$  & $91.33\PM3.75$  & $89.00\PM2.49$ \\
		 &10 & (5)&   $\textbf{92.67}\PM3.64$ & $90.40\PM3.90$ & $91.78\PM3.87$ &   $\textbf{93.20}\PM1.87$  & $91.20\PM1.71$  & $92.20\PM1.83$  \\ 
		 &20 & (10)&  $\textbf{92.65}\PM3.74$ & $91.28\PM4.36$ & $88.20\PM4.15$ &   $\textbf{92.80}\PM3.38$  & $91.80\PM3.96$  & $90.60\PM4.01$ \\ \midrule
                                                  
		 \multirow{4}{*}{0.3}
		 &0 & (0)&   $\textbf{90.98}\PM3.66$ & $90.66\PM3.82$ & $86.22\PM3.86$ &    -      &      -  &     - \\
		 &5  & (3)&  $\textbf{90.85}\PM3.69$ & $90.68\PM3.82$ & $86.13\PM3.78$ &   $\textbf{91.00}\PM2.16$  & $90.33\PM3.06$  & $86.33\PM2.49$\\
		 &10 & (5)&  $\textbf{90.58}\PM3.73$ & $90.40\PM3.95$ & $86.13\PM4.00$ &   $\textbf{91.60}\PM1.85$  & $91.20\PM1.82$  & $88.00\PM1.96$\\ 
		 &20 & (10)& $\textbf{90.70}\PM3.80$ & $90.20\PM4.07$ & $86.05\PM4.12$ &   $\textbf{91.30}\PM3.41$  & $90.60\PM3.80$  & $88.40\PM4.01$\\ \midrule
                                                  
		 \multirow{4}{*}{0.4}                    
		 &0  & (0)  &  $\textbf{90.98}\PM3.66$ & $90.66\PM3.82$ & $86.22\PM3.86$ &  -     &     -  &     - \\
		 &5  & (3)  &  $\textbf{90.85}\PM3.70$ & $90.66\PM3.95$ & $86.34\PM3.82$ & $\textbf{91.00}\PM2.16$ & $90.67\PM2.31$ & $86.33\PM2.39$ \\
		 &10 & (5)  &  $\textbf{90.58}\PM3.74$ & $90.40\PM4.00$ & $86.13\PM3.79$ & $\textbf{91.60}\PM1.85$ & $91.20\PM1.79$ & $88.00\PM1.78$ \\ 
		 &20 & (10) &  $\textbf{90.70}\PM3.81$ & $90.20\PM4.12$ & $86.03\PM4.06$ & $\textbf{91.30}\PM3.40$ & $90.60\PM4.01$ & $88.40\PM4.02$ \\ \bottomrule
	\end{tabular}      
}                     
 \vspace{-1.5em}
	\label{tab:mitigation_selfish_variable_mnist}
 \vspace{-0.1in}
\end{table}

\section{Conclusion}\label{sec:Conclusions}

We introduced a novel notion of \emph{selfish clients} who can deviate the overall \ac{fl} training in their favor.
From the server perspective, we proposed a robust aggregation strategy, RFL-Self, to mitigate the impact of these clients on the global model.
With rigorous analysis, we established that selfish clients can severely affect the training process and potentially deviate it to no convergence point.
By recovering true updates of selfish clients, the RFL-Self offered a strong robust aggregation strategy against selfishness.
By conducting extensive empirical analysis using two benchmark datasets with varying levels of selfishness, we observed that RFL-Self can handle the selfishness without degrading the model accuracy for normal clients and it is superior to other standard strategies like downscaling and median.
In the future, we plan to investigate adaptive selfishness, collusion among selfish clients, and the impact of selfish clients on fairness.

\bibliography{references}

\clearpage

\appendix

\section*{Appendix A: Experiments with 5 clients}

To assess the impact of selfish behavior and the effectiveness of RFL-Self in scenarios with few participating entities (such  as data silos), we perform the experimental evaluations on an FL setup with 5 clients. 
It is worth noting that, with so few clients, the selfish clients are the only ones holding data pertaining to their classes and it is thus crucial that any mitigation strategy manages to give good performance to all the clients.
Moreover, as only 5 clients are considered, having 2 selfish clients in the system already represents a 40\% concentration of misbehaving clients.

\subsection*{A1: Impact of selfishness on model performance}\label{subsubsec:Impact-of-selfish-clients-on-model-performance}

\begin{figure}[ht]
	\centering
	\includegraphics[width=0.95\linewidth, height=4.4cm]{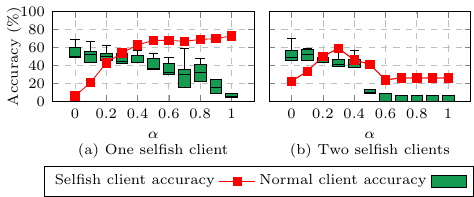}
     \vspace{0.1in}
	\caption{Test accuracy of the global model varying $\alpha$ with one and two selfish clients on the CIFAR-10 dataset.} 
     \vspace{0.1in}
	\label{fig:5-clients_no_mitigation_cifar}
\end{figure}

\begin{figure}[ht]
	\centering
	\includegraphics[width=0.95\linewidth, height=4.4cm]{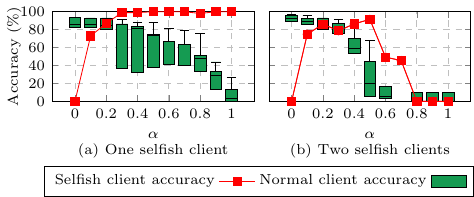}
     \vspace{0.1in}
	\caption{Test accuracy of the global model varying $\alpha$ with one and two selfish clients on the MNIST dataset.} 
     \vspace{0.1in}
	\label{fig:5-clients_no_mitigation_mnist}
\end{figure}

Figures~\ref{fig:5-clients_no_mitigation_cifar} and \ref{fig:5-clients_no_mitigation_mnist} show that, with few clients, a selfish client can more easily improve its performance at the expense of other clients.
Unsurprisingly, if $\alpha < \frac{1}{k}$, the selfish clients experience poor performances, as the global model receives no information pertaining to the classes in the selfish clients' datasets.

\begin{figure}[tb]
	\centering
	\includegraphics[width=0.9\linewidth, height=7cm]{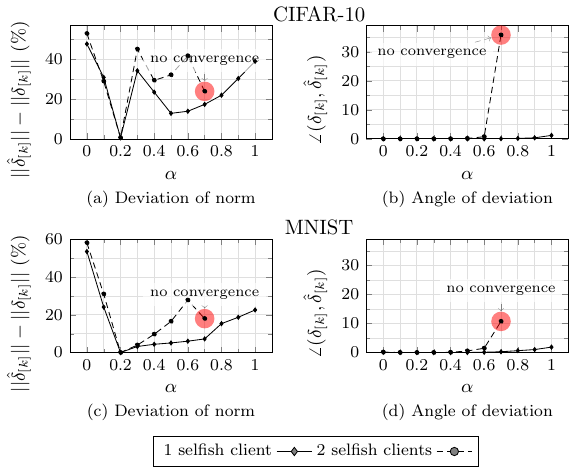}
     \vspace{0.1in}
	\caption{Deviation of the global update when all clients are normal to the one with selfish clients.}
     \vspace{0.1in}
	\label{fig:resilience_5}
\end{figure}

As in the setup with 50 clients, the presence of a second selfish client introduces instability and harms the convergence of the model (Figure~\ref{fig:resilience_5}).

\subsection*{A2: Performance of RFL-Self}

\begin{figure}[tb]
	\centering
		\includegraphics[width=0.9\linewidth, height=7cm]{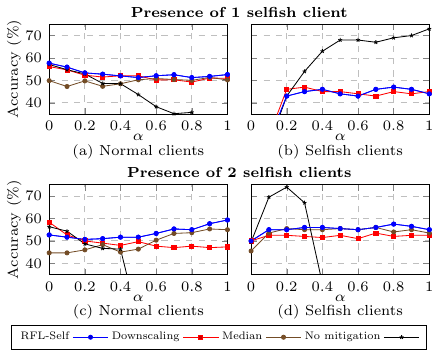}
     \vspace{0.1in}
	\caption{Performance of the robust aggregation strategies under an \ac{fl} setup with 5 clients using CIFAR-10 dataset.}
     \vspace{0.1in}
	\label{fig:mitigation_alpha_variable_cifar_5}
\end{figure}

On the CIFAR-10 dataset, the results in Figure~\ref{fig:mitigation_alpha_variable_cifar_5} are roughly similar to those reported for 50 clients.
An interesting point to notice is that, in the case of two selfish clients, the median, albeit underperforming RFL-self, is a preferable alternative to the downscaling strategy, yielding better accuracy for normal clients as $\alpha$ increases.

\begin{figure}[tb]
	\centering
		\includegraphics[width=0.9\linewidth, height=7cm]{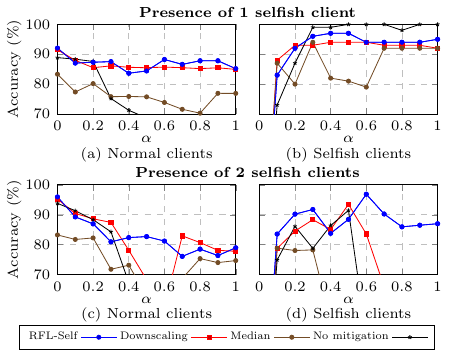}
     \vspace{0.1in}
	\caption{Performance of the robust aggregation strategies under an \ac{fl} setup with 5 clients using MNIST dataset.}
     \vspace{0.2in}
	\label{fig:mitigation_alpha_variable_mnist_5}
\end{figure}

The experiments on the MNIST dataset, reported in Figure~\ref{fig:mitigation_alpha_variable_mnist_5}, are more surprising.
Indeed, mitigation strategies other than RFL-self incur sudden severe drops in performance, for both selfish and normal clients, and are thus unreliable.
On the other hand, RFL-self is more stable and does not exhibit such abrupt variations in accuracy with varying $\alpha$.

\section*{Appendix B: Estimation of $k$}

The soundness of the proposed solution relies on the following assumptions:
\begin{enumerate}
	\item The average update of the normal clients $\bar\delta_{[k]\setminus \{s\}}$ is approximately equal between consecutive communication rounds $\bar\delta_{[k]\setminus \{s\}}^{t} \approx \bar\delta_{[k]\setminus \{s\}}^{t-1}$.
	\item The number of clients $k$ is fixed at the beginning of the FL process and does not change over time.
\end{enumerate}

The two assumptions make it possible for the selfish client to estimate the number of clients $k$ in the FL system and their average update vector $\bar\delta_{[k]\setminus \{s\}}$ by observing the past global update vectors.
Specifically, by taking into consideration the past $n$ global update vectors and the known selfish update vector $\hat\delta_{s}$, the selfish client can estimate the number of clients $k$ and the average update vector of the normal clients $\bar\delta_{[k]\setminus \{s\}}$ by solving the optimization problem in Eq.~\eqref{eq:delta_k}.
That is, finding the number of clients $k$ and the average update vector of the normal clients $\bar\delta_{[k]\setminus \{s\}}$ that minimizes the difference between the past global update vectors and the estimated global update vectors.

\begin{equation}\label{eq:delta_k}
	\arg\min_{\delta, k} \sum_{i=1}^n ||\bar\delta_{[k]}^{t-i}\cdot {k} - \hat\delta_{s}^{t-i} - \delta\cdot (k-1)||
\end{equation}

We chose to minimize the sum of the absolute differences between the past global update vectors and the estimated global update vectors because it is more robust to outliers than the sum of the higher powers of the differences.

\begin{algorithm}
	\begin{algorithmic}[1]
		\renewcommand{\algorithmicrequire}{\textbf{Input:}}
		\renewcommand{\algorithmicensure}{\textbf{Output:}}
		\REQUIRE $\bar\delta_{[k]}^{t-1}$, $\bar\delta_{[k]}^{t-2}$, $\hat\delta_{s}^{t-1}$, $\hat\delta_{s}^{t-2}$, $n$
		\ENSURE  $\hat{k}$, $\hat\delta_{[k]\setminus \{s\}}^{t}$

		\STATE $\hat\delta_{[k]\setminus \{s\}}^{t} \gets \textsc{geomed}(\bar\delta_{[k]}^{t-1} , \bar\delta_{[k]}^{t-2})$
		\WHILE {$\hat{k}$ does not converge}
		\STATE $\hat{k} \gets \arg\min_{k} \sum_{i=1}^n ||\bar\delta_{[k]}^{t-i}\cdot k - \hat\delta_{s}^{t-i} - \hat\delta_{[k]\setminus \{s\}}^{t}\cdot (k-1)||$
		\STATE $\hat\delta_{[k]\setminus \{s\}}^{t} \gets \arg\min_{\delta} \sum_{i=1}^n ||\bar\delta_{[k]}^{t-i}\cdot \hat{k} - \hat\delta_{s}^{t-i} - \delta\cdot (\hat{k}-1)||$
		\ENDWHILE
		\STATE \textbf{return} $\hat{k}$, $\hat\delta_{[k]\setminus \{s\}}^{t}$
	\end{algorithmic} 
	\caption{Estimation of the number of FL participants}
	\label{alg:Estimation-of-the-number-of-FL-participants}
\end{algorithm}

\begin{figure}[tb]
	\centering
     \vspace{0.1in}
	\includegraphics[width=\linewidth]{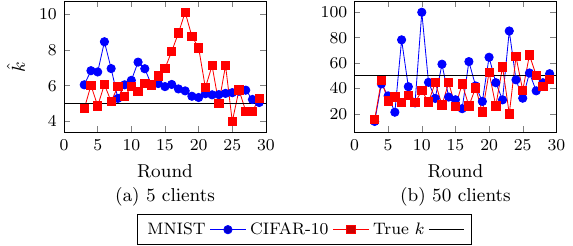}
     \vspace{0.1in}
	\caption{Estimation of $k$ across rounds for CIFAR-10 and MNIST. The estimation converges to the correct value.}
     \vspace{0.2in}
	\label{fig:k_estim}
\end{figure}

Algorithm~\ref{alg:Estimation-of-the-number-of-FL-participants} performs alternating minimization to estimate the number of FL participants and the average update attributed to normal clients as per Eq.~\eqref{eq:delta_k}.
We consider only the last two updates, i.e., $t-1$ and $t-2$, as the more updates are considered, the less accurate Assumption 1 becomes.
Line 3 of the algorithm can solved through Brent's method~\cite{brent1971algorithm}.
Line 4, instead, does not have an easily computable analytical solution.
Indeed, given two vectors, the point minimizing the sum of the $L_1$ distances from the two vectors is the geometric median of the two vectors, for which there is no explicit formula or exact algorithm.
Instead, we use a modified version of Weiszfeld's algorithm~\cite{pillutla2019robust}, which is a numerically stable iterative algorithm that converges to the geometric median in asymptotically linear time.
Thus, the geometric mean of $\bar\delta_{[k]}^{t-1}\cdot k - \hat\delta_{s}^{t-1}$ and $\bar\delta_{[k]}^{t-2}\cdot k\bar\delta_{s}^{t-2}$ is equal to $\delta\cdot (k-1)$, with $\delta$ satisfying the optimization problem in Line 4.
These two steps are repeated until convergence, i.e., until $\hat{k}$ does not change anymore.
As shown in Figure~\ref{fig:k_estim}, the estimation of $k$ converges to the correct value for both CIFAR-10 and MNIST datasets.

\begin{figure}[tb]
	\centering
	\includegraphics[width=\linewidth]{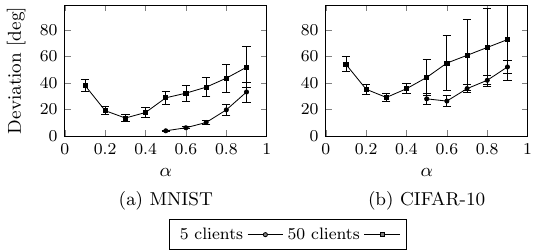}
     \vspace{0.1in}
	\caption{Angle between estimated and real average update vector across rounds for CIFAR-10 and MNIST. The estimation is relatively close to the real value.}
     \vspace{0.1in}
	\label{fig:deviation}
\end{figure}

As shown in Fig.~\ref{fig:deviation}, the median estimation error on the angle between the estimated and the real average update vector is 28.01 degrees for CIFAR-10 and 22.75 degrees for MNIST.
Considering that in high-dimensional spaces, random vectors are almost orthogonal to each other, this level of error is acceptable and indicates the effectiveness of the algorithm.
Indeed, the correlation coefficient between the estimated and the real average update vector reaches above 0.97 for both datasets.

\end{document}